\documentclass{article}
\usepackage{spconf,amsmath,graphicx}

\title{Brain Tumor Image Retrieval via Multitask Learning}

\name{
    Maxim Pisov\textsuperscript{* 1, 2} \qquad
    Gleb Makarchuk\thanks{\textsuperscript{*} Equal contribution}\textsuperscript{* 3}  \qquad
    Valery Kostjuchenko \textsuperscript{4}  \qquad
}
\secondlinename{
    Alexandra Dalechina \textsuperscript{4}  \qquad
    Andrey Golanov \textsuperscript{5} \qquad
    Mikhail Belyaev\textsuperscript{3, 1}
}

\address{
    Kharkevich Institute for Information Transmission Problems, Moscow, Russia \\
    Moscow Institute of Physics and Technology, Moscow, Russia \\
    Skolkovo Institute of Science and Technology, Moscow, Russia \\
    Moscow Gamma-Knife Center, Moscow, Russia \\
    Burdenko Neurosurgery Institute, Moscow, Russia \\
}

\begin{document}
%
\maketitle
\begin{abstract}
    Classification-based image retrieval systems are built by training convolutional neural networks (CNNs) on a relevant classification problem and using the distance in the resulting feature space as a similarity metric. However, in practical applications, it is often desirable to have representations which take into account several aspects of the data (e.g., brain tumor type and its localization). In our work, we extend the classification-based approach with multitask learning: we train a CNN on brain MRI scans with heterogeneous labels and implement a corresponding tumor image retrieval system. We validate our approach on brain tumor data which contains information about tumor types, shapes and localization. We show that our method allows us to build representations that contain more relevant information about tumors than single-task classification-based approaches.
\end{abstract}

\section{Introduction}

In connection with the success of the development of deep neural networks, the performance of machine learning algorithms in image analysis tasks has increased \cite{cnn_advances}. This progress had a positive impact on the analysis of medical images, where convolutional networks are now used to automate a variety of time-consuming clinical tasks or to enhance fundamental medical research \cite{2017_survey_dl_in_medicine}.
An important area of research is 3D brain MRI analysis by 3D CNNs. 
For example, in \cite{korolev2017residual}  the authors invistigated a simple 3D generalization of classical 2D CNNs for brain MRI classification.
In \cite{deepmedic} a 3D-convolutional network is proposed to address the problem of human brain MRI images segmentation. A more comprehensive analysis of deep learning applications to medical imaging classification and segmentation can be found in \cite{deep_learning_in_medical_detection}.

In our work, we focus on building a retrieval system for brain tumors. Such a system would help doctors to predict the development of diseases and facilitate radiosurgical treatment planning based on analysis of similar cases.
Previous related works can be split into three categories.  Multi-atlas medical image segmentation is a classical area of research \cite{konukoglu2013neighbourhood} which remains popular despite of increasing usage of CNNs for this task. The key drawback of this approach is time spending as multiple image-to-atlas registrations are needed. To increase processing speed, hashing forests are used to approximate nearest neighbor search to perform retrieval of similar atlases only \cite{katouzian2018hashing}. 

One of the most desired feature of any medical image retrieval system is the ability to extract semantically meaningful features from an image. 
There are two fundamentally different ways to build such a representation: one can create features manually using domain knowledge or an algorithm can be trained using labeled data. The former approach is still frequently used in medical imaging as sizes of datasets are still small. Such methods are based on heavy usage of classical preprocessing piplenes to extract a set of features. For example,  authors of \cite{faria2015content}
parcelled brain images into 211 regions and extracted morphometric features like ROI volumes or thickness to build a retrieval system. A similar approach was used in \cite{trojachanec2018longitudinal} where authors manually introduce a set of longitudinal features on top of morphometric ones.

Finally, the third set of related methods are based on CNNs. Deep convolutional networks are so effective due to, in part, their ability to automatically build an informative low dimensional representation of images space \cite{wan2014deep}. 
For example, in \cite{siamese_based_retrieval} authors train a Siamese-based CNN with a contrastive loss from comparing images of eyes for the presence of retinopathy. 
For  MR images such approach is frequently used as well.  Authors of \cite{qayyum2017medical} built a system to retrieve medical images using 24-organ classification problem for CNN training. \cite{shah_retrieval_using_cnn_and_hashing_forests} high-dimensional representations of prostate MRI images are built with the help of CNNs and then hashing forests are used to reduce the dimensionality of representations.

However, defining the similarity between two given tumors is an ill-posed and very challenging task, because there are multiple characteristics that must be taken into account, like type, shape (i.e., segmentation masks) and localization in the brain. To address this problem,  we utilize all these labels by training the network via multitask learning, simultaneously solving different supervised learning problems. It allows us to build representations that contain more relevant information about the tumors. 

\section{Method}
The main idea behind the method is to construct a function that maps the bounding box containing a tumor to a fixed-dimensional space, and then to treat the distance in the resulting space as a measure of similarity between the tumors.

To do so, we will assume that each brain MR image $x$  has a corresponding heterogeneous set of metadata: a segmentation mask $y^s$ and various labels for each tumor $y_t^p, t \in T, p \in P$ such as tumor type, localization e.t.c.
Here $T$ is the set of tumors in the brain $x$ and $P$ is the set of labels available for a given tumor.

\begin{figure*}
\begin{center}
  \includegraphics[width=0.9\linewidth]{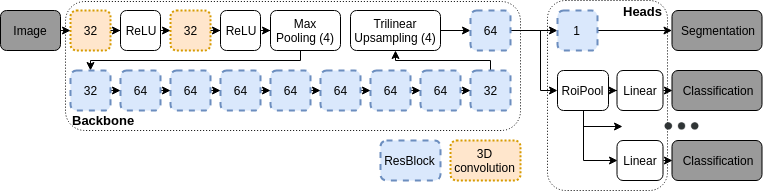}
  \caption{Network architecture. The number in each convolution and ResBlock cell represents the corresponding number of output channels. Downsampling (MaxPool) and upsampling are performed with a factor of 4.}
  \label{fig:architecture}
\end{center}
\end{figure*}

We build the mapping by training a neural network to solve multiple segmentation and classification tasks simultaneously.
The network's architecture (Fig. \ref{fig:architecture}) consists of two parts: a backbone CNN used to automatically extract features, and multiple heads each aimed at solving a particular task.

Since we want the resulting tumor representation to be as informative as possible, the head architectures are designed to be very simple, in order to "motivate" the backbone to generate a comprehensive feature map.

\subsection{Backbone}
\begin{figure}
    \begin{center}
      \includegraphics[width=0.9\linewidth]{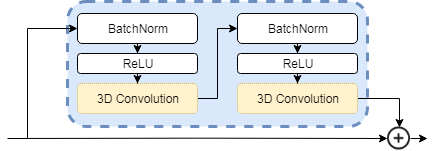}
      \caption{Schematic representation of a ResBlock}
      \label{fig:resblock}
    \end{center}
\end{figure}

The left side of Fig.\ref{fig:architecture} shows our backbone architecture: it is a relatively simple and fast ResNet-like \cite{resnet} network adapted for semantic segmentation. The first 2 convolutions are used to prepare the input image for the subsequent residual blocks (ResBlocks, Fig.\ref{fig:resblock}). We also apply downsampling and upsampling by a factor of 4 near the network's input and output respectively in order to decrease the amount of required memory and computation time.

The convolutions and ResBlocks have a kernel of size $3\times 3\times 3$ and padding $1\times 1\times 1$ which effectively leaves the input spatial shape unchanged after such operations, thus for an input image of shape $x\times y \times z$ the output feature map will have the shape $64\times x\times y \times z$.

\subsection{Network's heads}
Because the segmentation map is a global characteristic of the entire image, for the segmentation task we apply to the entire feature map the head which consists of a single ResBlock. (Fig. \ref{fig:resblock} shows the structure of a ResBlock.)

\begin{figure}
    \begin{center}
      \includegraphics[width=0.8\linewidth]{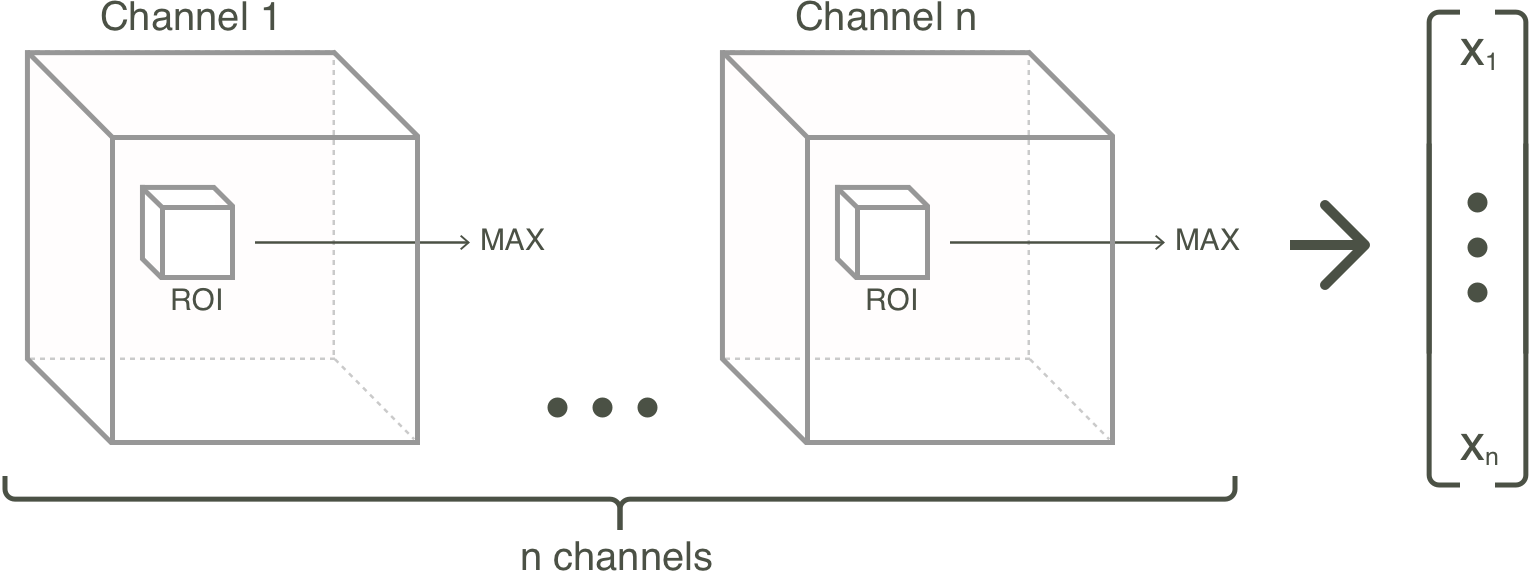}
      \caption{Schematic representation of RoiPool}
      \label{fig:roipool}
    \end{center}
    \vspace*{-6mm} 
\end{figure}

The tumor type and localization, however, is a property of a given image area containing a tumor, thus is a local characteristic. That's why for classification tasks we use a variation of the RoiPool block described in \cite{fast_rcnn}, followed by a single linear layer.
In our implementation of RoiPool we take a spatial slice corresponding to the tumor's bounding box and apply global max pooling to it (Fig. \ref{fig:roipool}). As a consequence, the output dimensionality is equal to the number of channels in the feature map generated by the backbone.

Note that RoiPool is essential for our method because our goal is to obtain fixed-size representations for tumors of different volumes. 

\subsection{Optimization criterion}
For each head, we use cross-entropy as the loss function (binary or multiclass, depending on the task being solved).
The final loss being optimized is a weighted sum:
\begin{equation}
    L_{total} = \lambda_s L(\tilde{y}^s, y^s) + \sum\limits_{p \in P} \lambda_p \sum\limits_{t \in T} L(\tilde{y}_t^p, y_t^p)
\end{equation}
where $L(\cdot, \cdot)$ is either binary or multiclass cross entropy, $y^s$ is the true segmentation mask, $y_t^p$ is the label for the tumor $t$ and task $p$, $\tilde{y}^s, \tilde{y}_t^p$ are the corresponding predictions and $\lambda_s, \lambda_p$ - the corresponding weights.

Note that in case of missing metadata for a given tumor and task, the corresponding term $L(\tilde{y}_t^p, y_t^p)$ can be simply omitted from the final loss.

\section{Experiments}
\subsection{Setup}
We train the network for 120 epochs with 200 batches (of size 2) per epoch. Because of technical limitations, instead of feeding entire images into the network, we use patches of size $120 \times 120 \times 120$. The patches are picked at random with the sole condition that each selected patch entirely contains a tumor.

In our experiments, we used stochastic gradient descent with Nesterov momentum. In order to avoid the train loss from plateauing we chose a decreasing learning rate policy: the initial learning rate - 0.1 was decreased by a factor of 10 at the 90th and 105th epochs.
The weights in the final loss were chosen during the baseline selection stage based on the model's performance: we ended up using $\lambda_s = 1$ for segmentation and $\lambda_p = 10^{-3}$ for classification. 

For all our experiments we used a random split (80\%/20\%) of the data with a fixed random seed. The networks were trained on 80\% of the data and the other 20\% were used for validation.

\subsection{Data}
\begin{table}[b]
\centering
\begin{tabular}{@{}lllll@{}}
                & Metastasis \hphantom & Meningioma \hphantom & Schwannoma \\ 
\# of images \hphantom & 399     & 339        & 251 \\
\# of tumors \hphantom & 1636    & 414        & 258 \\
\end{tabular}
\caption{Number of images and tumors in the dataset}
\label{dataset_description}
\end{table}

The data for our experiments was provided by a radiosurgical center that conducts operations on about 400 patients every year.
The dataset consists of 989 MRI images in the T1c modality with spatial pixel size $0.94 \times 0.94 \times 1 mm^3$ and typical image shapes of $200 \times 230 \times 170$, see details in Tab. \ref{dataset_description}. To avoid overfitting, we use only one MRI per patient. 

For each image a set of heterogeneous metadata is available: a binary mask of cancerous tissues and a set of labels (in rare cases some of them are missing) including lesion type (metastasis, meningioma, schwannoma), anatomical area such as \textit{Region frontalis} or \textit{Cerebellum} (in total we have 11 classes) and various localization: left/right hemisphere, front/rear and upper/lower.

\subsection{Validation}
Because the notion of tumors similarity is not well-formalized, the validation process is a very complicated task. In order to assess the quality of our approach we make the assumption that similar tumors resemble according to various measurable criteria: e.g. linear size (or volume), type, localization etc. We then evaluate how the distance between the tumors' embeddings correlates with the resemblance of these criteria.

\section{Results}
\begin{figure}
  \centering
  \includegraphics[width=\linewidth]{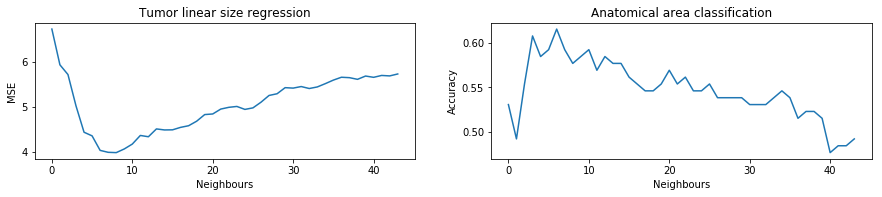}
  \caption{\label{fig:neighbours} KNN performance (accuracy for classification and RMSE(mm) for linear size regression) for classification and regression.}
\end{figure}

Fig.\ref{fig:neighbours} shows the performance of K-Nearest Neighbours models on various tasks depending on the number of neighbours. Note that the performance starts to decrease for $K > 5$, this suggests that the distance between the tumors representations correlates with their similarity. For further experiments we report KNN performances for $K = 5$.

Tab. \ref{tab:channels} shows that representations of higher dimensionality expectantly contain more information about some relevant characteristics. We couldn't go beyond 96 due to technical limitations. As a trade-off between accuracy and training time, in the rest of our experiments we use 64 channels.

\begin{table} 
    \includegraphics[width=\linewidth]{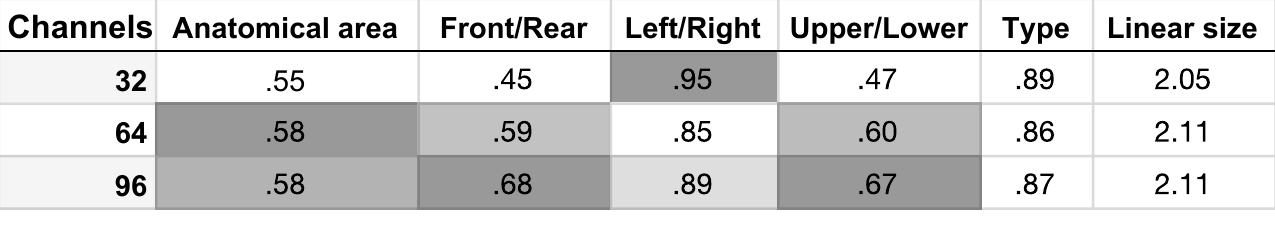}
    \caption{KNN performance (accuracy for classification and RMSE(mm) for linear size regression) depending on the number of channels.\label{tab:channels}}
\end{table}
 
\begin{table} 
    \centering
    \includegraphics[width=\linewidth]{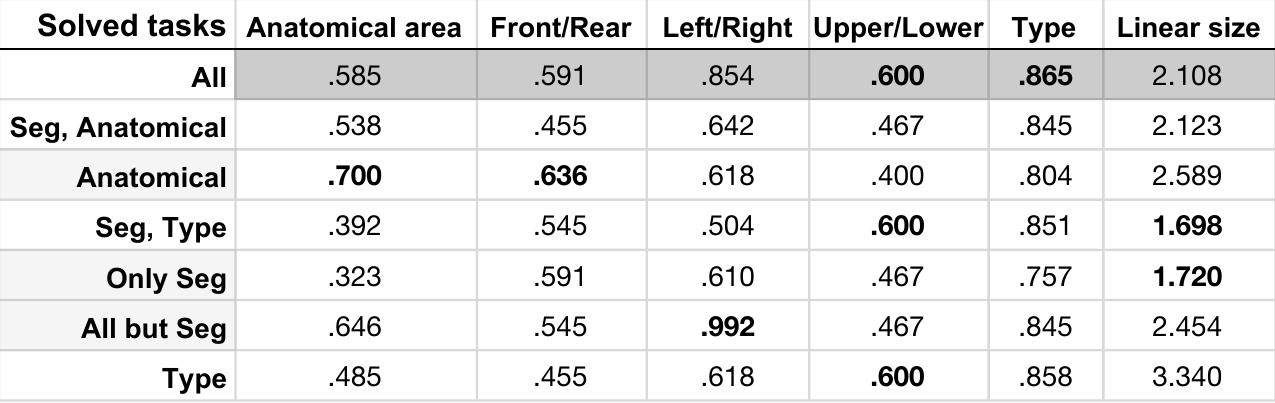}
    \caption{
        KNN performance (accuracy for classification and RMSE(mm) for linear size regression) depending on the tasks that the network was trained on. \label{tab:tasks}
    }
\end{table} 
 
Next, we analyze the feasibility of the multitask approach: we trained several networks excluding some tasks from the loss function. The results in Tab.\ref{tab:tasks} show that the multitask approach has a relatively balanced performance compared to the networks trained to solve a single problem, like tumor type classification, or tumors segmentation. This suggests that the network is able to aggregate the additional heterogeneous information contained in tumor labels.

\begin{figure}
  \centering
  \includegraphics[width=\linewidth]{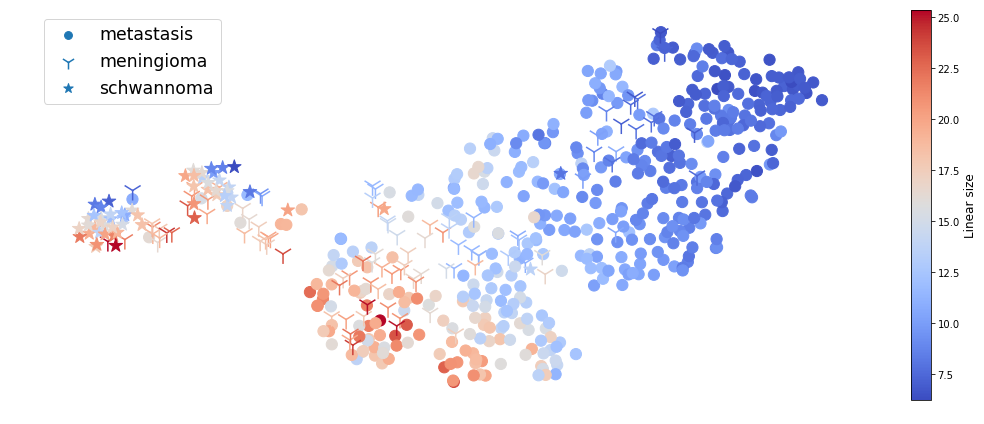}
  \caption{\label{fig:representations} Tumor representations after applying t-SNE. The color represents tumor linear sizes, the markers - tumor types.}
\end{figure}

We also visually analyze the representations by decreasing their dimensionality to 2D using using t-SNE. Fig. \ref{fig:representations} shows that the representations cluster by type and size. Note the two schwannoma clusters - each cluster corresponds to a brain hemisphere.

In is worth noting that in order to perform the search, our system requires a bounding box of the tumor, in a real world setting the provided bounding box might not be very accurate. In order to we analyze the robustness of our system to such inaccuracies, we added significant distortions to bounding boxes like scaling and translation, so that along each dimension: 
\begin{equation}
\begin{split}
    log_2(ScaleFactor) \sim \mathcal{N}(\mu=0, \sigma=1/3) \\ 
    Translation \sim BoxShape * \mathcal{N}(\mu=0, \sigma=1/10)
\end{split}
\end{equation}
We then took the first model from Tab.\ref{tab:tasks} (which saw only accurate boxes during training) and generated representations on the test set based on inaccurate boxes.
Tab.\ref{tab:distortions} shows that performance insignificantly declines on some tasks when the boxes are distorted, which means that the system is still able to find relevant tumors even with severely inaccurate bounding boxes.

\begin{table} 
    \centering
    \includegraphics[width=\linewidth]{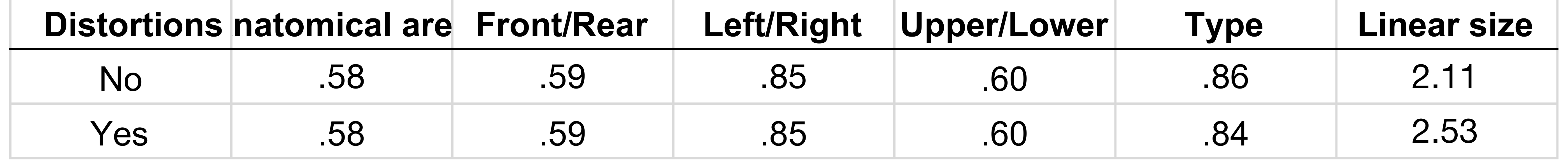}
    \caption{KNN performance (accuracy for classification and RMSE(mm) for linear size regression) with and without bounding box distortions.\label{tab:distortions}
    }
\end{table}

\begin{figure} 
    \centering
    \includegraphics[width=0.8\linewidth]{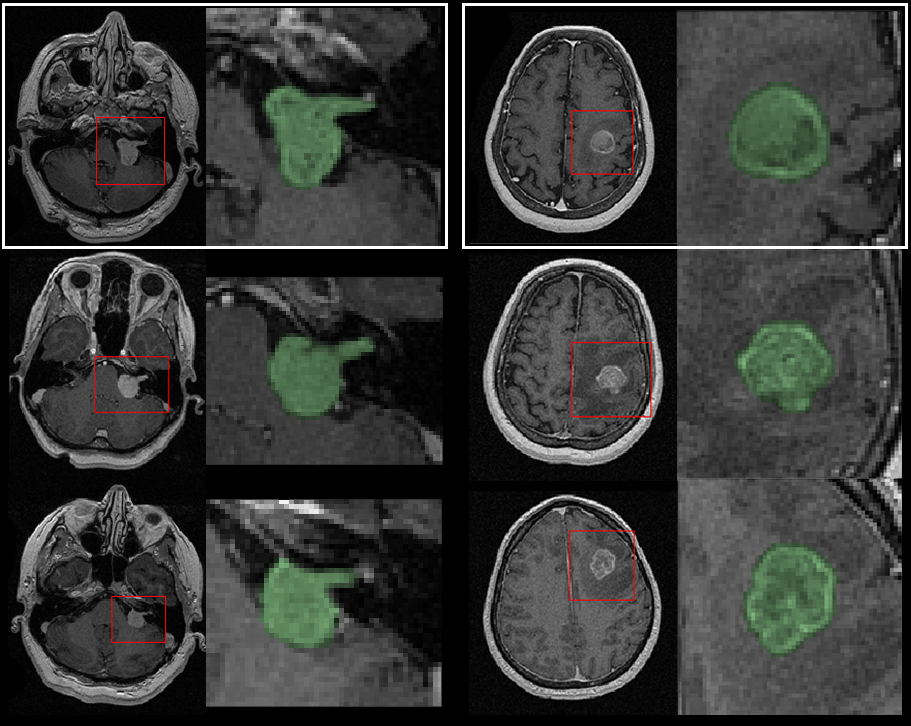}
    \caption{Retrieval examples: a reference schwannoma (top left) and metastasis (top right) and two tumors retrieved by the system for each case.\label{fig:retrieval_example}
    }
\end{figure}

Finally we show some tumor image retrieval examples (Fig.\ref{fig:retrieval_example}) for a randomly picked schwannoma and metastasis. Each example shows two tumors proposed by the system: note the similarities such as localization and shape (the resulting tumors are from different patients).

\section{Conclusion}
We developed a retrieval system that searches for similar brain tumors. The system is based on multitask learning which takes into account heterogeneous metadata available for each dataset entry. 

We demonstrated that the distance between tumor representations generated by our method correlates with various measurable tumor characteristics such as type, shape and localization. Also, we showed that the system can be used in the real-world setting by proving that the system performance remains almost unchanged even after significant bounding box distortion.

\bibliographystyle{IEEEbib}
\bibliography{main}

\end{document}